\documentclass[10pt,twocolumn,letterpaper]{article}

\usepackage{cvpr}              

\usepackage{graphicx}
\usepackage{amsmath}
\usepackage{amssymb}
\usepackage{booktabs}
\usepackage{bbm}
\usepackage{algorithm}
\usepackage{algpseudocode}
\usepackage{multirow}
\usepackage[accsupp]{axessibility}

\usepackage[pagebackref,breaklinks,colorlinks]{hyperref}

\usepackage[capitalize]{cleveref}
\crefname{section}{Sec.}{Secs.}
\Crefname{section}{Section}{Sections}
\Crefname{table}{Table}{Tables}
\crefname{table}{Tab.}{Tabs.}

\def\cvprPaperID{8063} 
\def\confName{CVPR}
\def\confYear{2022}

\begin{document}

\title{Improving Adversarial Transferability via Neuron Attribution-Based Attacks}

\author{Jianping Zhang$ ^{1} $ \qquad
Weibin Wu$ ^{2} $\thanks{Corresponding author.} \qquad
Jen-tse Huang$ ^{1} $ \qquad
Yizhan Huang$ ^{1} $ \qquad
\\
Wenxuan Wang$ ^{1} $ \qquad
Yuxin Su$ ^{2} $ \qquad
Michael R. Lyu$ ^{1} $
\\
$ ^{1} $Department of Computer Science and Engineering, The Chinese University of Hong Kong
\\
$ ^{2} $School of Software Engineering, Sun Yat-sen University
\\
{\tt\small \{jpzhang, jthuang, yzhuang9, wxwang, lyu\}@cse.cuhk.edu.hk, \{wuwb36, suyx35\}@mail.sysu.edu.cn}
}

\maketitle

\begin{abstract}
   Deep neural networks (DNNs) are known to be vulnerable to adversarial examples. It is thus imperative to devise effective attack algorithms to identify the deficiencies of DNNs beforehand in security-sensitive applications. To efficiently tackle the black-box setting where the target model's particulars are unknown, feature-level transfer-based attacks propose to contaminate the intermediate feature outputs of local models, and then directly employ the crafted adversarial samples to attack the target model. Due to the transferability of features, feature-level attacks have shown promise in synthesizing more transferable adversarial samples. However, existing feature-level attacks generally employ inaccurate neuron importance estimations, which deteriorates their transferability. To overcome such pitfalls, in this paper, we propose the Neuron Attribution-based Attack (NAA), which conducts feature-level attacks with more accurate neuron importance estimations. Specifically, we first completely attribute a model's output to each neuron in a middle layer. We then derive an approximation scheme of neuron attribution to tremendously reduce the computation overhead. Finally, we weight neurons based on their attribution results and launch feature-level attacks. Extensive experiments confirm the superiority of our approach to the state-of-the-art benchmarks. Our code is available at: \href{https://github.com/jpzhang1810/NAA}{https://github.com/jpzhang1810/NAA} .
\end{abstract}


\section{Introduction}
\label{sec:intro}

\begin{figure}[htbp]
\centering
\begin{minipage}[t]{0.155\textwidth}
\centering
\includegraphics[width=2.5cm]{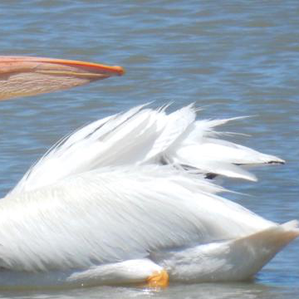}
\subcaption*{Benign Image}
\end{minipage}
\begin{minipage}[t]{0.155\textwidth}
\centering
\includegraphics[width=2.5cm]{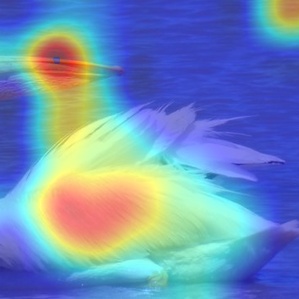}
\subcaption*{Source Model Attention}
\end{minipage}
\begin{minipage}[t]{0.155\textwidth}
\centering
\includegraphics[width=2.5cm]{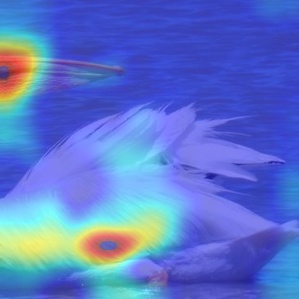}
\subcaption*{Target Model Attention}
\end{minipage}
\begin{minipage}[t]{0.155\textwidth}
\centering
\includegraphics[width=2.5cm]{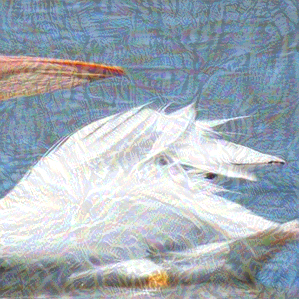}
\subcaption*{Adversarial Image}
\end{minipage}
\begin{minipage}[t]{0.155\textwidth}
\centering
\includegraphics[width=2.5cm]{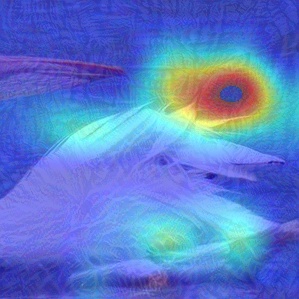}
\subcaption*{Source Model Attention}
\end{minipage}
\begin{minipage}[t]{0.155\textwidth}
\centering
\includegraphics[width=2.5cm]{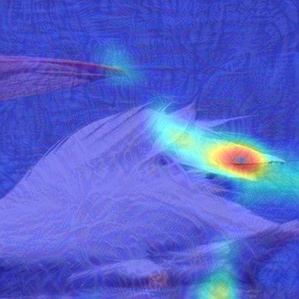}
\subcaption*{Target Model Attention}
\end{minipage}
\vspace*{-3mm}
\caption{Visualization of model attentions on both the benign image and adversarial image generated by our method. The attentions of both the source model and target model change dramatically on the adversarial image compared with the benign image.}
\label{intro}
\end{figure}

Deep neural networks (DNNs) have been deployed in many safety-critical real-world applications, such as autonomous driving and medical diagnosis. However, recent research shows that DNNs are vulnerable to adversarial attacks \cite{szegedy2013intriguing}, which add human-imperceptible perturbations to clean images to mislead DNNs. It is thus imperative to devise effective attack algorithms to identify the deficiencies of DNNs beforehand, which serves as the first step to improve their robustness.

There are generally two categories of adversarial attacks: white-box and black-box attacks. Attackers under the white-box setting can fetch the structures and parameters of the target models to craft adversarial examples. In contrast, under the black-box setting, attackers have no access to the model structure and parameters. In real-world applications, the DNN models are generally deployed in the black-box situation. Therefore, we focus on black-box attacks in this work.

Black-box attacks can be roughly divided into query-based and transfer-based schemes. Query-based methods approximate the gradient information by queries \cite{brendel2017decision, uesato2018adversarial, ilyas2018black} to generate adversarial examples. However, query-based methods are impractical since large quantities of queries are not allowed in reality. As a result, researchers turn to efficient transfer-bases attacks \cite{dong2018boosting,lin2019nesterov,xie2019improving,gao2020patch,dong2019evading}, which employ white-box attacks to attack a local surrogate model, and then directly transfer the resultant adversarial samples to the target model. Instead of directly manipulating the local model's final output, feature-level transfer-based attacks propose to destroy the intermediate feature maps of local models. Since the most critical features are shared among different DNN models \cite{naseer2018task, ganeshan2019fda}, feature-level transfer-based attacks have shown promise in relieving the overfitting issue and synthesizing more transferable adversarial samples \cite{wang2021feature}.

However, existing feature-level transfer-based attacks still have limited transferability due to the reliance on inappropriate neuron importance measures. NRDM \cite{naseer2018task} views all neurons as important neurons and tries to maximize the distortion of neuron activation after attacks. However, in a middle layer, there are positive and negative features that promote and suppress the correct prediction of models, respectively. As a result, maximizing the feature distortion destroys positive and negative features at the same time, while the negative features should be enhanced for generating adversarial samples. FDA \cite{ganeshan2019fda} differentiates the polarity of neuron importance by mean activation values. Unfortunately, it still attaches the same importance to all neurons except for their signs. FIA \cite{wang2021feature} measures the neuron importance by the multiplication of neuron activation and back-propagated gradients. However, the back-propagated gradient on the original input suffers from the problem of saturation \cite{dhamdhere2018important}.

To address the drawbacks of existing feature-level transfer-based attacks, in this paper, we propose Neuron Attribution-based Attack (NAA), which conducts feature-level attacks based on more accurate neuron importance measures. Specifically, inspired by the neuron attribution method \cite{dhamdhere2018important}, we first attempt to completely attribute a model's output to each neuron. It ensures that our neuron attribution results possess a good property of completeness that the sum of all the neuron attribution results equals to the output value. Consequently, the attribution results can accurately reflect the attribution of each neuron to the output, taking into consideration both the polarity and magnitude of neuron importance. From Figure \ref{intro}, our attack method finds the important features on the mouse instead of the lake, which means our method can accurately find more important features that can craft more transferable adversarial examples. However, directly utilizing neuron attribution method is intractable due to extensive computation consumption. We then devise an approximation approach to conduct neuron attribution to tremendously reduce the computation cost. Finally, we weight each neuron according to their attribution results, and endeavor to minimize the weighted feature output. Comprehensive experiments confirm the superiority of our method. Our contributions are:\vspace{-1.5ex}

\begin{itemize}
    \item We deploy neuron attribution method to better measure neuron importance when launching feature-level attacks. We further devise an approximation for neuron attribution, which largely reduces the time consumption and promotes the attack efficiency.\vspace{-1ex}
    \item Based on the proposed neuron importance measure, we devise a novel feature-level attack, Neuron Attribution-based Attack (NAA), to overcome the drawbacks of existing feature-level attacks and improve the transferability of adversarial examples.\vspace{-1ex}
    \item Comprehensive experiments validate the effectiveness and efficiency of our method. We can achieve state-of-the-art performance on attacking both undefended and defended models.
\end{itemize}

\section{Related Work}


\subsection{Adversarial Attacks}

Adversarial attacks generally have two categories: white-box attack and black-box attack. Attackers can access the information of victim models like model structure and parameters under the white-box setting, while the attackers fail to fetch the information of the victim models in the black-box setting. 
Many methods adopt the gradient information of the victim model to launch adversarial attacks under the white-box setting, like Fast Gradient Sign Method (FGSM) \cite{goodfellow2014explaining}, Iterative Fast Gradient Sign Method (I-FGSM) \cite{kurakin2016adversarial}, Project Gradient Descent (PGD) \cite{madry2017towards}, and Carlini and Wagner Attack (C\&W) \cite{carlini2017towards}. However, white box attacks are unrealistic in real applications because the model structure and parameters are hidden from the users.

Therefore, black-box adversarial attacks are of more significance. In this paper, we mainly focus on the transfer-based adversarial attack. Transferability is a phenomenon in which adversarial examples crafted by the source model have the ability to mislead other models. Therefore, we utilize the transferability of adversarial examples to launch the black-box adversarial attack. Many works are proposed to improve the transferability of adversarial examples by advanced gradient, like Momentum Iterative Method (MIM) \cite{dong2018boosting} and Nesterov Iterative Method (NIM) \cite{lin2019nesterov}. In addition to the modification of the gradient, input transformation methods adopt the image transformation methods on the input image to generate more transferable adversarial examples, like Diverse Input Method (DIM) \cite{xie2019improving}, Patch-wise Iterative Method (PIM) \cite{gao2020patch}, Translation Invariant Method (TIM) \cite{dong2019evading}, and Scale Invariant Method (SIM). Input transformation methods can be composed with any other adversarial attack methods to further improve the transferability of adversarial examples.

In addition to crafting adversarial examples on the output layer, some works pay attention to the internal layers. Transferable Adversarial Perturbations (TAP) \cite{zhou2018transferable} observes that maximizing the distance between the adversarial example and benign image on the feature map enhances the transferability of adversarial examples. NRDM \cite{naseer2018task} follows the same idea and generates high-strength adversarial examples that are transferable across different network architectures and different vision tasks (image segmentation, classification and object detection). Intermediate Level Attack (ILA) \cite{huang2019enhancing} fine-tunes existing adversarial examples by increasing the perturbation on a target layer from the source model to further enhance the transferability. Feature Disruptive Attack (FDA) \cite{ganeshan2019fda} introduces a new attack method motivated by corrupting features at the target layer. Although FDA differentiates the polarity of neuron importance by mean activation values, previous methods treat all neurons as important neurons. Feature Importance-aware Attack (FIA) \cite{wang2021feature} measures the neuron importance by the multiplication of activation and the back-propagated on the target layer. However, the back-propagated gradient on the original input suffers from the problem of saturation \cite{dhamdhere2018important}, which fail to measure the real importance. Though previous methods generate transferable adversarial examples, their inappropriate measurement can not represent the real effect of each neuron to the output. Our approach utilizes neuron attribution as the measurement to reflect the real influence on the output. We thus deploy the neuron attribution to craft adversarial examples. We suppress the weighted sum of neuron attributions to destroy the positive features and promote the negative features at the same time. Utilizing the neuron attribution paves an explainable and more transferable way to do feature-level adversarial attacks.

\subsection{Adversarial Defenses}

Adversarial defenses are of great importance to alleviate the threats of adversarial attacks. Adversarial defenses generally have two categories: adversarial training and denoising. Adversarial training is a simple but effective way to defend the adversarial attacks \cite{tramer2017ensemble, kurakin2016adversarial} because DNNs are data-driven. Consequently, retraining the models by adding the adversarial examples into the training data improves the model robustness dramatically \cite{goodfellow2014explaining}. Additionally, ensemble adversarial training injects the adversarial examples transferred from several models to defend transfer-based attacks \cite{kurakin2016adversarial}. While, denoising filters out the adversarial perturbations by pre-processing mechanisms before feeding the data into the models. The models can correctly classify the rectified input images without the loss of performance. The state-of-the-art defense methods include utilizing a high-level representation guided denoiser \cite{liao2018defense}, random resizing and padding \cite{xie2017mitigating}, JPEG based defensive compression framework \cite{liu2019feature}, compression module \cite{jia2019comdefend}, and randomized smoothing \cite{cohen2019certified}. In this paper, we exploit these state-of-the-art defenses to validate the superior of our attack against advanced defended models.

\section{Approach}

Feature-level attacks follow the observation that the DNN models share similar features in their receptive fields \cite{wu2020boosting} and craft adversarial examples by destroying the positive features or enlarging the negative features. Therefore, the adversarial examples generated by feature-level attacks inherit the highly transferable features which can mislead other DNN models. The key point to craft feature-level attacks is to find a proper way of measuring the importance of each neuron for representing feature patterns. In this section, we introduce a measurement of neuron importance, namely neuron attribution. Then we propose an approximation for neuron attribution which reduces the computation cost greatly. Finally, we propose our approach, Neuron Attribution-based Attack via the estimation of neuron importance.

We denote the benign image to be $x$ and its corresponding true label as $z$. Then we assume a classification model $F(\cdot)$ where $F(x)$ represents the output with the input image $x$. Furthermore, $y$ denotes the activation values of the $y$-th layer while $y_j$ denotes the activation value of the $j$-th neuron on this feature map. We aim to craft the adversarial example $x^{adv}$ by injecting imperceptible perturbation on the input image to mislead the model while satisfying the constraints $\left\| x - x^{adv} \right\|_{p} < \epsilon$. The $\left\| \cdot \right\|_{p}$ represents the $p$-norm distance and we follow the previous works \cite{dong2018boosting,wang2021feature} to focus on the $L_{\infty}$-norm distance in this paper.

Inspired by \cite{sundararajan2017axiomatic} and \cite{dhamdhere2018important}, we define the attribution of input image $x$ (with $N \times N$ pixels) with respect to a baseline image $x'$ as
\begin{equation} \label{eq1}
\begin{split}
    A & := \sum_{i=1}^{N^2} (x_i - x_i') \int_{0}^{1} \frac{\partial F}{\partial x_i}(x' + \alpha(x - x')) \ d\alpha, \\
\end{split}
\end{equation}
where $\frac{\partial F}{\partial x_i}(\cdot)$ denotes the partial derivative of $F$ to the $i$-th pixel. Equation \ref{eq1} is a path integration of the gradient of $F$ along the straight line given by $(x' + \alpha(x - x'))$. Applying the fundamental theorem of calculus for path integrals, we can show that $A \approx F(x)$ as long as $F(x') \approx 0$. In practice, a black image (i.e., $x'=\mathbf{0}$) serves well as this baseline.

\begin{algorithm}[t]
\caption{Neuron Attribution-based Attack}\label{alg1}
\begin{algorithmic}
\Require classifier $F$, and target layer $y$
\Require positive and negative transformation function $f_p(\cdot)$ and $f_n(\cdot)$, and hyperparameter $\gamma$
\Require benign input $x$ with label $z$
\Require perturbation budget $\epsilon$ and iteration number $T$
\Require baseline image $x'$ and integrated step $n$
\State $\alpha = \frac{\epsilon}{T}$, $x_0^{adv} = x$, $IA = \mathbf{0}$, $g_0 = \mathbf{0}$, $\mu = 1$
\For{$m = 1 \gets n$}
\State $IA = IA + \nabla_{y(x'+ \frac{m}{n}(x - x'))}F(x'+ \frac{m}{n}(x - x'))$
\EndFor
\State $IA = IA / n$
\For{$t = 0 \gets T-1$}
\State $A_y = (y - y') \cdot IA$
\State $WA_y = \sum_{{\begin{subarray}{c} A_{y_j} \geq 0 \\ y_j \in y \end{subarray}}} f_p(A_{y_j}) - \gamma \cdot \sum_{{\begin{subarray}{c} A_{y_j} < 0 \\ y_j \in y \end{subarray}}} f_n(-A_{y_j})$
\State $g_{t+1} = \mu \cdot g_t + \frac{\nabla_{x} WA_y}{\left \| \nabla_{x} WA_y \right\|_{1}}$
\State $x_{t+1}^{adv} = Clip_{x}^{\epsilon} \{x_{t+1}^{adv} - \alpha \cdot sgn(g_{t+1})\}$
\EndFor
\end{algorithmic}
\end{algorithm}

Then we can attribute the attribution $A$ to each neuron in a certain layer $y$. With denoting $x' + \alpha(x - x') = x_\alpha$, the attribution of the neuron $y_j$ is
\begin{equation} \label{eq2}
\begin{split}
    A_{y_j} & = \sum_{i=1}^{N^2} (x_i - x_i') \int_{0}^{1} \frac{\partial F}{\partial y_j}(y(x_\alpha)) \frac{\partial y_j}{\partial x_i}(x_\alpha) \ d\alpha. \notag\\
\end{split}
\end{equation}
Note that $\sum_{y_j \in y} A_{y_j} = A$ always holds no matter which layer we choose. Therefore, neuron attribution reflects the real influence of each neuron to the output. To compute the integral in practice, we sample $n$ virtual images along the straight line and use the Riemann sum to approximate the integral. And after changing the order of the summation, we have
\vspace*{-2mm}
\begin{equation} \label{eq3}
\begin{split}
    A_{y_j} & \approx \frac{1}{n} \sum_{m=1}^{n} \bigg(\frac{\partial F}{\partial y_j}(y(x_m))\bigg) \bigg(\sum_{i=1}^{N^2} (x_i - x_i') \frac{\partial y_j}{\partial x_i}(x_m)\bigg), \\
\end{split}
\end{equation}
where $x_m = x' + \frac{m}{n}(x - x')$ are the virtual images.

As shown in Equation \ref{eq3}, we have to compute the gradient $\frac{\partial y_j}{\partial x_i}$ for each neuron. Consequently, the computation cost is extremely high considering the number of neurons in the DNNs. To reduce the computation time, we make a simple assumption to simplify Equation \ref{eq3}. To begin with, $\frac{\partial F}{\partial y_j}(y(x_m))$ is the gradient of $F(x)$ to the neuron $y_j$, related to the latter layers after $y$. Meanwhile, $\sum_{i=1}^{N^2} (x_i - x_i') \frac{\partial y_j}{\partial x_i}(x_m)$ is the sum of the gradient of $y_j$ to each pixel $x_i$, related to the former layers of the network. Given the fact that the former part and latter part are independent in most traditional DNN models, we assume that the two parts are linearly independent, i.e., the two gradient sequences should have zero covariance.

Note that given two sequences $a_i$ and $b_i$ with zero covariance, we have $\sum_{1}^{n} (a_i - \bar{a_i})(b_i - \bar{b_i}) = 0$, where $\bar{(\cdot)}$ is the mean of the sequence. After the expansion, we have $\sum_{1}^{n} a_i\cdot b_i = \frac{1}{n}\sum_{1}^{n}a_i \cdot \sum_{1}^{n}b_i$. Regarding the components in the two big brackets in Equation \ref{eq3} as $a$ and $b$ respectively, we have
\begin{equation} \label{eq4}
\begin{split}
    A_{y_j} & \approx \frac{1}{n} \sum_{m=1}^{n} \frac{\partial F}{\partial y_j}(y(x_m)) \frac{1}{n} \sum_{m=1}^{n} \sum_{i=1}^{N^2} (x_i - x_i') \frac{\partial y_j}{\partial x_i}(x_m). \notag
\end{split}
\end{equation}
By applying the fundamental theorem of calculus for path integrals, we have $\frac{1}{n} \sum_{m=1}^{n} \sum_{i=1}^{N^2} (x_i - x_i') \frac{\partial y_j}{\partial x_i}(x_m) = (y_j - y_j')$ where $y_j'$ is the activation value of the neuron when the input is a black image. With denoting $y_j - y_j'$ as $\Delta y_j$ and $\frac{1}{n} \sum_{m=1}^{n} \frac{\partial F}{\partial y_j}(y(x_m))$ as Integrated Attention $IA(y_j)$, we have a simpler form of $A_{y_j} \approx \Delta y_j \cdot IA(y_j)$. The name of $IA(y_j)$ reflects the integration of the gradient along the straight line from the baseline image to the input with attention to the neuron $y_j$.

\begin{table*}
\centering
\begin{tabular}{|c|c|ccccccccc|} 
\hline
Model & Attack & Inc-v3 & Inc-v4 & IncRes-v2 & Res-v2 &\begin{tabular}[c]{@{}c@{}}Inc-v3\\${\text{adv}}$\end{tabular}  & \begin{tabular}[c]{@{}c@{}}IncRes-v2\\${\text{adv}}$\end{tabular}  &\begin{tabular}[c]{@{}c@{}}Inc-v3\\${\text{ens3}}$\end{tabular} & \begin{tabular}[c]{@{}c@{}}Inc-v3\\${\text{ens4}}$\end{tabular} & \begin{tabular}[c]{@{}c@{}}IncRes-v2\\${\text{ens3}}$\end{tabular}  \\ 
\hline
\multirow{5}{*}{Inc-v3} & MIM & \bf 100.0 & 41.1 & 39.9 & 32.7 & 22.9 & 19.3 & 16.0 & 16.5 & 8.1  \\
 & NRDM & 90.9 & 61.3 & 53.9 & 50.8 & 26.6 & 18.7 & 9.8 & 10.3 & 5.1 \\
 & FDA & 81.3 & 42.9 & 36.0 & 35.4 & 19.3 & 12.2 & 8.9 & 6.4 & 2.3 \\
 & FIA & 98.3 & 83.2 & 79.1 & 71.6 & 53.3 & 50.8 & 36.1 & 37.0 & 20.0 \\
 & NAA & 98.1 & \bf 85.0 & \bf 82.4 & \bf 77.1 & \bf 61.5 & \bf 62.7 & \bf 50.5 & \bf 50.8 & \bf 31.5 \\
\hline 
\multirow{5}{*}{Inc-v4} & MIM & 58.2 & \bf 99.7 & 45.5 & 38.6 & 23.8 & 21.2 & 18.7 & 18.5 & 8.9\\
 & NRDM & 78.2 & 97.4 & 61.9 & 61.9 & 26.1 & 26.0 & 17.7 & 15.7 & 5.6 \\
 & FDA & 84.8 & 99.6 & 71.9 & 68.7 & 27.9 & 25.9 & 18.4 & 17.2 & 7.3 \\
 & FIA & 84.1 & 95.7 & 78.6 & 72.0 & 45.3 & 47.3 & 38.0 & 37.2 & 19.4 \\
 & NAA & \bf 86.0 & 96.5 & \bf 81.0 & \bf 75.5 & \bf 52.4 & \bf 56.0 & \bf 50.5 & \bf 49.4 & \bf 30.8 \\
\hline 
\multirow{5}{*}{IncRes-v2} & MIM & 59.5 & 51.0 & \bf 99.2 & 42.3 & 25.3 & 30.9 & 21.8 & 23.7 & 12.7 \\
 & NRDM & 71.0 & 66.8 & 77.3 & 57.8 & 34.3 & 29.6 & 16.2 & 23.8 & 19.4 \\
 & FDA & 69.3 & 67.7 & 78.3 & 56.3 & 36.4 & 29.8 & 16.2 & 22.3 & 17.9 \\
 & FIA & 81.6 & 77.1 & 88.7 & 71.0 & 63.8 & 65.0 & 49.8 & 46.6 & 34.1 \\
 & NAA & \bf 82.4 & \bf 78.0 & 93.0 & \bf 74.4 & \bf 64.9 & \bf 67.1 & \bf 60.0 & \bf 56.7 & \bf 47.5 \\
\hline
\multirow{5}{*}{Res-v2} & MIM & 54.1 & 47.5 & 45.3 & \bf 99.4 & 26.4 & 25.1 & 24.2 & 25.3 & 12.4 \\
 & NRDM & 73.6 & 70.9 & 58.8 & 90.4 & 39.5 & 30.3 & 23.7 & 19.9 & 9.5\\
 & FDA & 83.9 & 84.1 & 73.9 & 89.1 & 51.2 & 42.9 & 27.9 & 23.6 & 11.5\\
 & FIA & 83.0 & 81.6 & 78.4 & 98.9 & 58.2 & 58.2 & 49.1 & 44.9 & 29.3 \\
 & NAA & \bf 85.9 & \bf 85.0 & \bf 83.6 & 98.2 & \bf 66.1 & \bf 69.8 & \bf 61.6 & \bf 59.2 & \bf 46.7 \\
\hline
\end{tabular}
\vspace*{-3mm}
\caption{The attack success rates (\%) on four undefended models and five adversarially trained models by various momentum optimization based attacks. The adversarial examples are crafted on Inc-v3, Inc-v4, IncRes-v2, and Res-v2, respectively. The best result is in bold.}
\label{table1}
\end{table*}

All in all, we approximate the attribution of each neuron on the feature map by the multiplication of relative activation $\Delta y_j$ and Integrated Attention on the neuron $IA(y_j)$. The computation complexity of neuron attribution is $\mathcal{O}(H*W*C)$, where $H$ is the height of the target layer, $W$ is the width of the target layer, and $C$ is the channel number of the target layer. While our computation complexity is $\mathcal{O}(1)$. Note that we only need one gradient operation in each integration step. Conversely, we have to take about nearly one million gradient operations in each step if we do not simplify the Equation \ref{eq3}. Hence, our approximation saves the computation time to a significant extent. Now, we demonstrate our proposed Neuron Attribution-based Attack (NAA). Since minimizing the total neuron attributions to the output can reduce the positive attributions and enlarges the negative attributions at the same time, we consider the attribution of all neurons in a same layer $y$ calculated by
\vspace*{-1mm}
\begin{equation} \label{eq5}
\begin{split}
    A_{y} = \sum_{y_j \in y} A_{y_j} = \sum_{y_j \in y} \Delta y_j \cdot IA(y_j) = (y - y') \cdot IA(y). \notag
\end{split}
\end{equation}
In consequence, useful features are suppressed and harmful features are amplified. To analyze the influence of the two kinds of features and figure out which one dominates the transferability of adversarial examples, we utilize a hyperparameter $\gamma$ to balance between the positive and negative attributions. Furthermore, we aim to distinguish the significant degree of neuron attributions with different values. For example, we investigate whether decreasing a large positive attribution neuron may benefit the attack more compared to increasing a small negative attribution neuron. To this end, we design multiple linear or non-linear transformation functions, namely $f_p(A_{y_j})$ for positive neuron attribution and $-f_n(-A_{y_j})$ for negative neuron attribution. Therefore, the Weighted Attribution $WA_y$ of all neurons on the target layer $y$ can be computed with
\begin{equation} \label{eq6}
\begin{split}
    WA_{y} & = \sum_{{\begin{subarray}{c} A_{y_j} \geq 0 \\ y_j \in y \end{subarray}}} f_p(A_{y_j}) - \gamma \cdot \sum_{{\begin{subarray}{c} A_{y_j} < 0 \\ y_j \in y \end{subarray}}} f_n(-A_{y_j}). \notag
\end{split}
\end{equation}
Minimizing $WA_y$ is better than minimizing $A_y$ directly in practice since $WA_y$ takes the neuron attribution polarity and value magnitude into consideration. Hence, the goal of our proposed NAA is formulated into solving the following constrained minimization problem:
\begin{equation}
    \min_{x^{adv}} \ \ WA_{y} \ \ \ \ s.t. \left \| x - x^{adv} \right \|_{\infty} < \epsilon. \notag
    \label{eq7}
\end{equation}
We deploy MIM \cite{dong2018boosting} to solve this constrained minimization problem. The whole process of running NAA algorithm is shown in Algorithm \ref{alg1}.

\section{Experiments}

In this section, we launch extensive experiments to evaluate the effectiveness of our proposed methods. We first clarify the setup of the experiments. After that, we illustrate the attacking results of our methods against competitive baseline methods under various experimental settings and state the attack effectiveness on advanced defense models. The experiment results demonstrate the effectiveness of our methods that further improve the transferability of adversarial examples compared with baseline methods. Furthermore, we analyze the positive and negative attribution transformation functions as well as the hyperparameter $\gamma$ to understand the significance of neuron attributions with different polarities and values. Finally, we present the ablation study on the target feature map layers and the hyperparameter $n$ in the Integrated Attention equation.

\begin{table*}
\centering
\begin{tabular}{|c|c|ccccccccc|} 
\hline
Model & Attack & Inc-v3 & Inc-v4 & IncRes-v2 & Res-v2 &\begin{tabular}[c]{@{}c@{}}Inc-v3\\${\text{adv}}$\end{tabular}  & \begin{tabular}[c]{@{}c@{}}IncRes-v2\\${\text{adv}}$\end{tabular}  &\begin{tabular}[c]{@{}c@{}}Inc-v3\\${\text{ens3}}$\end{tabular} & \begin{tabular}[c]{@{}c@{}}Inc-v3\\${\text{ens4}}$\end{tabular} & \begin{tabular}[c]{@{}c@{}}IncRes-v2\\${\text{ens3}}$\end{tabular}  \\ 
\hline
\multirow{5}{*}{Inc-v3} & MIM-PD & \bf 99.8 & 70.0 & 67.6 & 53.6 & 31.0 & 28.0 & 21.3 & 22.0 & 9.3  \\
& NRDM-PD & 87.3 & 66.7 & 62.8 & 59.5 & 29.7 & 22.9 & 12.2 & 18.6 & 13.6 \\
 & FDA-PD & 76.0 & 50.4 & 46.5 & 39.2 & 23.0 & 16.0 & 10.8 & 12.1 & 8.0 \\
 & FIA-PD & 98.7 & 87.2 & 86.1 & 80.1 & 59.8 & 57.1 & 38.5 & 37.3 & 21.5 \\
 & NAA-PD & 98.8 & \bf 89.4 & \bf 88.4 & \bf 83.6 & \bf 67.9 & \bf 68.6 & \bf 55.4 & \bf 55.6 & \bf 33.8 \\
\hline 
\multirow{5}{*}{Inc-v4} & MIM-PD & 81.4 & \bf 99.3 & 72.0 & 59.4 & 30.6 & 28.8 & 23.9 & 24.5 & 12.5\\
& NRDM-PD & 88.8 & 97.0 & 80.2 & 78.4 & 34.2 & 35.0 & 21.3 & 19.2 & 8.6 \\
 & FDA-PD & 91.4 & 99.2 & 87.1 & 82.2 & 36.6 & 38.0 & 21.9 & 20.9 & 9.1 \\
 & FIA-PD & 90.6 & 97.1 & 88.8 & 84.9 & 55.3 & 60.7 & 45.5 & 42.1 & 23.5 \\
 & NAA-PD & \bf 91.5 & 97.7 & \bf 89.7 & \bf 86.5 & \bf 61.3 & \bf 87.9 & \bf 55.4 & \bf 53.6 & \bf 34.4 \\
\hline 
\multirow{5}{*}{IncRes-v2} & MIM-PD & 80.6 & 76.5 & \bf 98.1 & 64.0 & 36.7 & 41.7 & 28.8 & 26.7 & 16.3 \\
 & NRDM-PD & 76.5 & 75.4 & 79.6 & 66.3 & 40.8 & 32.3 & 18.6 & 30.6 & 26.0 \\
 & FDA-PD & 78.6 & 76.0 & 80.3 & 66.3 & 41.2 & 35.6 & 17.4 & 29.9 & 25.3 \\
 & FIA-PD & 85.1 & 79.9 & 90.9 & 76.5 & 66.9 & 66.7 & 49.7 & 44.9 & 31.9 \\
 & NAA-PD & \bf 85.5 & \bf 82.5 & 93.9 & \bf 79.3 & \bf 69.4 & \bf 71.3 & \bf 61.9 & \bf 59.0 & \bf 48.3 \\
\hline
\multirow{5}{*}{Res-v2} & MIM-PD & 81.8 & 76.7 & 75.7 & \bf 99.4 & 42.0 & 44.5 & 36.3 & 34.3 & 18.1 \\
 & NRDM-PD & 60.6 & 55.9 & 50.0 & 87.2 & 26.2 & 18.2 & 13.8 & 14.5 & 5.9\\
 & FDA-PD & 64.7 & 60.1 & 56.5 & 92.1 & 28.7 & 21.5 & 13.6 & 15.5 & 7.1\\
 & FIA-PD & 90.0 & 88.4 & 87.9 & 98.7 & 71.0 & 69.7 & 58.3 & 53.9 & 34.6 \\
 & NAA-PD & \bf 92.0 & \bf 90.7 & \bf 90.3 & 98.7 & \bf 76.0 & \bf 78.9 & \bf 72.4 & \bf 68.0 & \bf 52.8 \\
\hline
\end{tabular}
\vspace*{-3mm}
\caption{The attack success rates (\%) on four undefended models and five adversarially trained models by various momentum optimization based attacks with input transformations (PIM and DIM). The adversarial examples are crafted on Inc-v3, Inc-v4, IncRes-v2, and Res-v2, respectively. The best result is in bold.}
\label{table2}
\end{table*}

\subsection{Experiment Setup}

We follow the protocol of the baseline method \cite{wang2021feature} to set up the experiments for a fair comparison to attack image classification models trained on ImageNet \cite{russakovsky2015imagenet}. ImageNet is also the most widely utilized benchmark task for transfer-based adversarial attacks \cite{carlini2019evaluating, kurakin2018adversarial, wu2021improving}. Here are the details of the experiment setup.

\textbf{Dataset}. We follow the dataset of the baseline method \cite{wang2021feature} by randomly sampling 1000 images of different categories from the ILSVRC 2012 validation set \cite{russakovsky2015imagenet}. We check that all of the attacking models are almost approaching 100\% classification success rate in this paper.

\textbf{Models}. We choose four representative models containing Inception-v3 (Inc-v3) \cite{szegedy2016rethinking}, Inception-v4 (Inc-v4) \cite{szegedy2017inception}, Inception-Resnet-v2 (IncRes-v2) \cite{szegedy2017inception} and Resnet-v2-152 (Res-v2) \cite{he2016deep, he2016identity} as the source model to craft adversarial examples. We consider undefended (normally trained) models and defended (adversarial training and advanced defense technique) models as the target models. For undefended models, we use the four source models as the target models. For defended models, we consider adversarial training and advanced defense models because adversarial training is a simple but effective technique \cite{madry2017towards} and advanced defense models are robust against black-box adversarial attacks.
We select five adversarially trained models: adversarially trained Inception-v3 (Inc-v3$_{\text{adv}}$), ensemble of three adversarially trained Inception-v3 models (Inc-v3$_{\text{ens3}}$), ensemble of four adversarially trained Inception-v3 models (Inc-v3$_{\text{ens4}}$), adversarially trained Inception-Resnet-v2 (IncRes-v2$_{\text{adv}}$) and ensemble of three adversarially trained Inception-Resnet-v2 models (IncRes-v2$_{\text{ens3}}$). We also select seven advanced defense methods covering random resizing and padding (R\&P) \cite{xie2017mitigating}, NIPS-r3\footnote{https://github.com/anlthms/nips-2017/tree/master/mmd}, feature distillation (FD) \cite{liu2019feature}, compression defense (ComDefend) \cite{jia2019comdefend}, and randomized smoothing (RS) \cite{cohen2019certified}, PGD-based adversarial training (PGD) \cite{salman2020adversarially}, and Fast adversarial training (Fast) \cite{wong2020fast}.

\textbf{Baseline Methods}. We choose the advanced gradient-based iterative adversarial attacks: MIM \cite{dong2018boosting} as our baseline, which we also utilize as an optimization method. Additionally, we select three feature-level adversarial attack methods: NRDM \cite{naseer2018task}, FDA \cite{ganeshan2019fda} and FIA \cite{wang2021feature} as our competitive baselines, where FIA is state-of-the-art. NRDM directly increases the difference between the original example and adversarial example on the target feature map. FDA utilizes mean activation to split the feature map into positive and negative activation then they suppress the positive activation and enhance the negative activation. FIA computes the average gradient of the input with random drop transformation \cite{srivastava2014dropout} as the attention and reduces the multiplication of the attention and activation on the target layer. We compare our approach with them in various settings to validate the effectiveness of our method. In addition, we integrate all the methods with two well-known input transformation methods: DIM \cite{xie2019improving} and PIM \cite{gao2020patch} to further validate the superiority of our method. We denote our method combined with input transformation methods as NAA-PD. The basic baseline method MIM combined with input transformation methods as MIM-PD. Furthermore, we denote other feature-level adversarial attacks combined with input transformation methods as NRDM-PD, FDA-PD, and FIA-PD.

\textbf{Evaluation}. The attack success rate is the ratio of the adversarial examples that successfully mislead the target model among all the generated adversarial examples. Therefore, we utilize the attack success rate on the target model by the crafted adversarial examples to evaluate the attacking performance.
\begin{table*}
\centering
\begin{tabular}{|c|cccccccc|} 
\hline
Attack & R\&P & NIPS-r3 & FD & ComDefend & RS & PGD & Fast & Average \\ 
\hline
MIM-PD  & 22.4 & 28.8 & 62.5 & 59.5 & 31.4 & 42.6 & 33.6 & 40.1 \\
NRDM-PD & 16.9 & 23.9 & 43.2 & 43.8 & 28.0 & 41.8 & 34.0 & 33.1 \\
FDA-PD  & 16.3 & 23.1 & 37.0 & 37.7 & 27.8 & 41.5 & 30.5 & 30.6 \\
FIA-PD  & 36.4 & 51.2 & 76.7 & 74.3 & 38.4 & 44.9 & 42.3 & 52.0 \\
NAA-PD  & \bf 46.8 & \bf 62.9 & \bf 83.2 & \bf 80.9 & \bf 40.4 & \bf 46.8 & \bf 43.9 & \bf 57.9 \\
\hline

\end{tabular}
\vspace*{-3mm}
\caption{The attack success rates (\%) of the adversarial examples on seven advanced defense mechanisms. The adversarial examples are generated on the Inc-v3 model. The best result is in bold.}
\label{table3}
\end{table*}

\textbf{Parameter}. For a fair comparison, we follow the parameter setting in \cite{wang2021feature} to set the maximum perturbation of $\epsilon = 16$ and the number of iteration $T = 10$, so the step length $\alpha = \frac{\epsilon}{T} = 1.6$. Furthermore, We set the decay factor $\mu = 1.0$ for all the baselines because all the baselines utilize the momentum method as the optimizer. For the input transformation methods, we set transformation probability to be 0.7 for DIM. we take the amplification factor to be 2.5 and kernel size to be 3 for PIM. For our own method, we follow \cite{smilkov2017smoothgrad} to implement the Integrated Attention and we choose the middle layer to be the target layer. Specifically, we select to attack Mixed\_5b for Inception-v3 (Inc-v3), Mixed\_5e for Inception-v4 (Inc-v4), Conv2d\_4a\_3x3 for Inception-Resnet-v2 (IncRes-v2) and the last layer of block2 for Resnet-v2-152 (Res-v2). To compare with the state-of-the-art baselines, we treat neuron attributions with different polarities and values equally. Therefore, we let $\gamma = 1$ and the transformation functions degrade to the linear functions.

\subsection{Attack Results}

\begin{figure}[t]
  \centering
  \centerline{\includegraphics[width=0.9\linewidth]{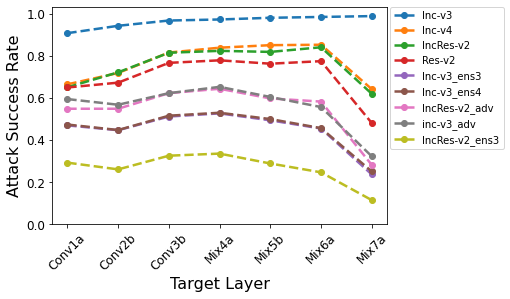}}
  \vspace*{-5mm}
  \caption{
  The  attack  success  rates  (\%)  of  NAA under different target layer settings.
  }
  \label{fig4}
  \vspace*{-3mm}
\end{figure}

In this section, we analyze the performance of our approach against the undefended models, adversarially trained models and models with advanced defenses respectively. Specifically, we attack a given source model and directly test the other different models by crafted adversarial examples, which is the black-box setting. We also test the adversarial examples on the source model itself in a white-box setting. 

We can see from Table \ref{table1}, our approach achieves nearly 100 percent attack accuracy under the white-box setting. Our method outperforms all the baselines in the black-box setting which illustrates the high transferability of our method. Although our method has a similar white-box attacking success rate with FIA and a little bit worse white-box attacking success rate than MIM, our method is more transferable with a high attack success rate under the black-box setting.

Then, we study the performance of our proposed attacking method against the adversarially trained models. As also shown in Table \ref{table1}, NAA outperforms all of the baselines under all the settings with a large margin of 10.5 \%, which validates our method has a strong attacking ability against adversarially trained models. Especially, our method has a similar white-box attack success rate with FIA and a worse white-box attack success rate than MIM, but our approach is more transferable. 

Furthermore, we compose all the attacking methods with input transformation methods: PIM and DIM to further improve the transferability as shown in Table \ref{table2}. Our approach combined with input transformation methods also outperforms all the baseline methods by a considerable margin of 10.7\% on average under the black-box setting, which further demonstrates the superiority of our method.

In addition, we assess the performance of our proposed NAA and other baseline attacks against the models with advanced defense mechanisms. We first take inc-v3 as the source model and generate adversarial examples for all the baseline methods with transformation inputs methods: PIM and DIM. Then we test the prediction accuracy of adversarial examples on advanced defended models as shown in Table \ref{table3}. Our proposed method achieves 57.9 \% attack success rate on average and surpasses all of the baselines more than a margin of 5.9\%, which shows a strong threat to state-of-the-art defense methods.

From the above experiments, our proposed method has more transferability compared with all of the baselines. We conclude the reasons why NAA has strong transferability are two-folded. First of all, the neuron attribution provides a simple but effective way to model the importance of neurons, which reflects the real attribution to the output. Furthermore, the independent assumption simplifies the representation of neuron attribution and improves the transferability of generated adversarial examples at the same time. To illustrate, we consider a simple scenario when we attack the target models that only change the later networks of the source model. If we assume the former networks and later networks of source models are related, the generated adversarial examples will overfit the source model. Transfer attack between the source model (Inc-v3) and source model variants (Inc-v4) as the target model can partially validate the reasons.

\subsection{Ablation Study}

In this section, we do ablation studies to analyze the three factors in our proposed NAA. The first factor is the target feature map layer to figure out which layer (shallow layer, middle layer or deep layer) is prone to craft transferable adversarial examples. The second factor is the integrated steps number $n$ to find its relationship with transferability. The last factor is the weighted attribution including $\gamma$, $f_p(\cdot)$ and $f_n(\cdot)$ to examine the importance of neuron attributions with different polarities and values. 

\begin{figure}[t]
  \centering
  \centerline{\includegraphics[width=0.9\linewidth]{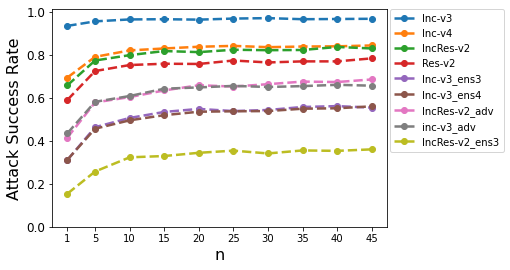}}
  \vspace*{-5mm}
  \caption{
  The  attack  success  rates  (\%)  of  NAA with different $ n $ values.
  }
  \label{fig5}
\end{figure}

\textbf{Target Layer}. We utilize NAA on different target layers to craft adversarial examples and observe the transferability. We choose Inc-v3 as the source model and different target layers based on the network structure stages. As shown in Figure \ref{fig4}, attacking the deep layers (Mix6a/Mix7a) achieves the best white-box attack performance. However, attacking middle layers (Mix4a/Mix5b) achieves the higher transferable performance compared with the shallow layers (Conv1a/Conv2b) and deep layers (Mix6a/Mix7a). We believe the shallow layers contain low-level features which exert less influence on the output. Similarly, the deep layers contain high-level features but the attack on the deep layers overfits the source model failing to craft transferable adversarial examples. As a result, attacking the middle-level features achieves the best performance.

\textbf{Integrated Steps}. We measure the transferability of adversarial examples generated from the Inc-v3 model by altering integrated steps. We observe from Figure \ref{fig5} that with the increase of integrated step, the transferability boosts. Although the performance is improved, the computation cost increases with $n$. In order to balance the performance and computation cost, we choose $n=30$ to achieve adequate performance.

\begin{figure}[t]
  \centering
  \centerline{\includegraphics[width=0.9\linewidth]{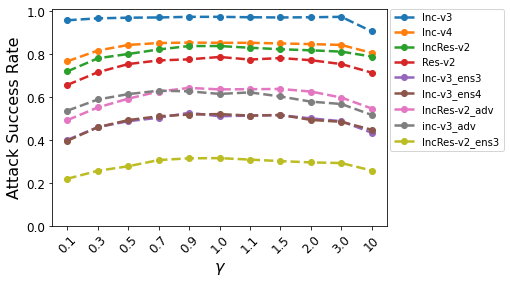}}
  \vspace*{-5mm}
  \caption{
  The  attack  success  rates  (\%)  of  NAA with different $\gamma$ values.
  }
  \label{fig6}
\end{figure}

\textbf{Weighted Attribution}. We study the neuron attribution from two sides: the polarity of neuron attribution and the value of neuron attribution to figure out the significance. We first analyze the importance between the positive attribution and negative attribution by altering the value of $\gamma$ in Equation \ref{eq5}. As shown in Figure \ref{fig6}, the attack success rate rises when we increase $\gamma$ and it decreases when $\gamma$ is greater than $1$. Therefore, $\gamma = 1$ achieves the best performance, which implies the negative attributions are as significant as positive attributions. Hence, positive attributions and negative attributions are equally important. 

\begin{figure}[t]
\centering
\begin{minipage}[t]{0.23\textwidth}
\centering
\includegraphics[width=4cm]{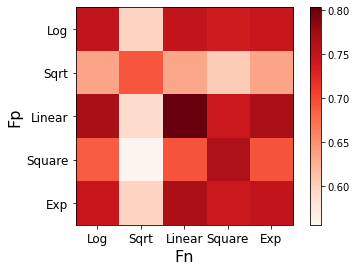}
\subcaption*{Undefended Models}
\end{minipage}
\begin{minipage}[t]{0.23\textwidth}
\centering
\includegraphics[width=4cm]{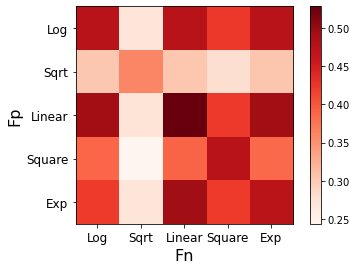}
\subcaption*{Adversarially Trained Models}
\end{minipage}
\vspace*{-3mm}
\caption{Heat map of the average attack success rate on undefended models and adversarially trained models under different combinations of transformation functions.}
\vspace*{-3mm}
\label{fig7}
\end{figure}

After that, we try different transformation functions to measure the neuron attributions with different values, like the exponential function that focuses more on the high value and the logarithm function which focuses more on the low value. We attack the Inc-v3 model to generate adversarial examples based on different combinations of $f_p(\cdot)$ and $f_n(\cdot)$ in Equation \ref{eq5}. We choose five transformation functions: logarithm function, square root function, linear function, square function and exponential function. As shown in Figure \ref{fig7}, the combination of linear functions has the best performance, which implies the attributions with different values have the same importance. All in all, we should treat all attributions with different polarities or values equally.

\section{Conclusion}

In this paper, we propose the Neuron Attribution-based Attack (NAA) to craft transferable adversarial examples. Specifically, we first employ neuron attribution to more accurately estimate the neuron importance. To reduce the computation time, we then derive an approximation scheme for neuron attribution. Finally, we minimize the weighted combination of the positive and negative neuron attribution values to generate adversarial samples. Experimental results corroborate that our method can outperform state-of-the-art baselines by a considerable margin. 

\section*{Acknowledgment}
The work described in this paper was supported by the key program of fundamental research from the Shenzhen Science and Technology Innovation Commission (No. JCYJ20200109113403826) and the Research Grants Council of the Hong Kong Special Administrative Region, China (CUHK 14210920 of the General Research Fund).

{\small
\bibliographystyle{ieee_fullname}
\bibliography{egbib}

\begin{thebibliography}{10}\itemsep=-1pt

\bibitem{brendel2017decision}
Wieland Brendel, Jonas Rauber, and Matthias Bethge.
\newblock Decision-based adversarial attacks: Reliable attacks against
  black-box machine learning models.
\newblock {\em arXiv preprint arXiv:1712.04248}, 2017.

\bibitem{carlini2019evaluating}
Nicholas Carlini, Anish Athalye, Nicolas Papernot, Wieland Brendel, Jonas
  Rauber, Dimitris Tsipras, Ian Goodfellow, Aleksander Madry, and Alexey
  Kurakin.
\newblock On evaluating adversarial robustness.
\newblock {\em arXiv preprint arXiv:1902.06705}, 2019.

\bibitem{carlini2017towards}
Nicholas Carlini and David Wagner.
\newblock Towards evaluating the robustness of neural networks.
\newblock In {\em 2017 ieee symposium on security and privacy (sp)}, pages
  39--57. IEEE, 2017.

\bibitem{cohen2019certified}
Jeremy Cohen, Elan Rosenfeld, and Zico Kolter.
\newblock Certified adversarial robustness via randomized smoothing.
\newblock In {\em International Conference on Machine Learning}, pages
  1310--1320. PMLR, 2019.

\bibitem{dhamdhere2018important}
Kedar Dhamdhere, Mukund Sundararajan, and Qiqi Yan.
\newblock How important is a neuron?
\newblock {\em arXiv preprint arXiv:1805.12233}, 2018.

\bibitem{dong2018boosting}
Yinpeng Dong, Fangzhou Liao, Tianyu Pang, Hang Su, Jun Zhu, Xiaolin Hu, and
  Jianguo Li.
\newblock Boosting adversarial attacks with momentum.
\newblock In {\em Proceedings of the IEEE conference on computer vision and
  pattern recognition}, pages 9185--9193, 2018.

\bibitem{dong2019evading}
Yinpeng Dong, Tianyu Pang, Hang Su, and Jun Zhu.
\newblock Evading defenses to transferable adversarial examples by
  translation-invariant attacks.
\newblock In {\em Proceedings of the IEEE/CVF Conference on Computer Vision and
  Pattern Recognition}, pages 4312--4321, 2019.

\bibitem{ganeshan2019fda}
Aditya Ganeshan, Vivek BS, and R~Venkatesh Babu.
\newblock Fda: Feature disruptive attack.
\newblock In {\em Proceedings of the IEEE/CVF International Conference on
  Computer Vision}, pages 8069--8079, 2019.

\bibitem{gao2020patch}
Lianli Gao, Qilong Zhang, Jingkuan Song, Xianglong Liu, and Heng~Tao Shen.
\newblock Patch-wise attack for fooling deep neural network.
\newblock In {\em European Conference on Computer Vision}, pages 307--322.
  Springer, 2020.

\bibitem{goodfellow2014explaining}
Ian~J Goodfellow, Jonathon Shlens, and Christian Szegedy.
\newblock Explaining and harnessing adversarial examples.
\newblock {\em arXiv preprint arXiv:1412.6572}, 2014.

\bibitem{he2016deep}
Kaiming He, Xiangyu Zhang, Shaoqing Ren, and Jian Sun.
\newblock Deep residual learning for image recognition.
\newblock In {\em Proceedings of the IEEE conference on computer vision and
  pattern recognition}, pages 770--778, 2016.

\bibitem{he2016identity}
Kaiming He, Xiangyu Zhang, Shaoqing Ren, and Jian Sun.
\newblock Identity mappings in deep residual networks.
\newblock In {\em European conference on computer vision}, pages 630--645.
  Springer, 2016.

\bibitem{huang2019enhancing}
Qian Huang, Isay Katsman, Horace He, Zeqi Gu, Serge Belongie, and Ser-Nam Lim.
\newblock Enhancing adversarial example transferability with an intermediate
  level attack.
\newblock In {\em Proceedings of the IEEE/CVF International Conference on
  Computer Vision}, pages 4733--4742, 2019.

\bibitem{ilyas2018black}
Andrew Ilyas, Logan Engstrom, Anish Athalye, and Jessy Lin.
\newblock Black-box adversarial attacks with limited queries and information.
\newblock In {\em International Conference on Machine Learning}, pages
  2137--2146. PMLR, 2018.

\bibitem{jia2019comdefend}
Xiaojun Jia, Xingxing Wei, Xiaochun Cao, and Hassan Foroosh.
\newblock Comdefend: An efficient image compression model to defend adversarial
  examples.
\newblock In {\em Proceedings of the IEEE/CVF Conference on Computer Vision and
  Pattern Recognition}, pages 6084--6092, 2019.

\bibitem{kurakin2018adversarial}
Alexey Kurakin, Ian Goodfellow, Samy Bengio, Yinpeng Dong, Fangzhou Liao, Ming
  Liang, Tianyu Pang, Jun Zhu, Xiaolin Hu, Cihang Xie, et~al.
\newblock Adversarial attacks and defences competition.
\newblock In {\em The NIPS'17 Competition: Building Intelligent Systems}, pages
  195--231. Springer, 2018.

\bibitem{kurakin2016adversarial}
Alexey Kurakin, Ian Goodfellow, Samy Bengio, et~al.
\newblock Adversarial examples in the physical world, 2016.

\bibitem{liao2018defense}
Fangzhou Liao, Ming Liang, Yinpeng Dong, Tianyu Pang, Xiaolin Hu, and Jun Zhu.
\newblock Defense against adversarial attacks using high-level representation
  guided denoiser.
\newblock In {\em Proceedings of the IEEE Conference on Computer Vision and
  Pattern Recognition}, pages 1778--1787, 2018.

\bibitem{lin2019nesterov}
Jiadong Lin, Chuanbiao Song, Kun He, Liwei Wang, and John~E Hopcroft.
\newblock Nesterov accelerated gradient and scale invariance for adversarial
  attacks.
\newblock {\em arXiv preprint arXiv:1908.06281}, 2019.

\bibitem{liu2019feature}
Zihao Liu, Qi Liu, Tao Liu, Nuo Xu, Xue Lin, Yanzhi Wang, and Wujie Wen.
\newblock Feature distillation: Dnn-oriented jpeg compression against
  adversarial examples.
\newblock In {\em 2019 IEEE/CVF Conference on Computer Vision and Pattern
  Recognition (CVPR)}, pages 860--868. IEEE, 2019.

\bibitem{madry2017towards}
Aleksander Madry, Aleksandar Makelov, Ludwig Schmidt, Dimitris Tsipras, and
  Adrian Vladu.
\newblock Towards deep learning models resistant to adversarial attacks.
\newblock {\em arXiv preprint arXiv:1706.06083}, 2017.

\bibitem{naseer2018task}
Muzammal Naseer, Salman~H Khan, Shafin Rahman, and Fatih Porikli.
\newblock Task-generalizable adversarial attack based on perceptual metric.
\newblock {\em arXiv preprint arXiv:1811.09020}, 2018.

\bibitem{russakovsky2015imagenet}
Olga Russakovsky, Jia Deng, Hao Su, Jonathan Krause, Sanjeev Satheesh, Sean Ma,
  Zhiheng Huang, Andrej Karpathy, Aditya Khosla, Michael Bernstein, et~al.
\newblock Imagenet large scale visual recognition challenge.
\newblock {\em International journal of computer vision}, 115(3):211--252,
  2015.

\bibitem{salman2020adversarially}
Hadi Salman, Andrew Ilyas, Logan Engstrom, Ashish Kapoor, and Aleksander Madry.
\newblock Do adversarially robust imagenet models transfer better?
\newblock {\em Advances in Neural Information Processing Systems},
  33:3533--3545, 2020.

\bibitem{smilkov2017smoothgrad}
Daniel Smilkov, Nikhil Thorat, Been Kim, Fernanda Vi{\'e}gas, and Martin
  Wattenberg.
\newblock Smoothgrad: removing noise by adding noise.
\newblock {\em arXiv preprint arXiv:1706.03825}, 2017.

\bibitem{srivastava2014dropout}
Nitish Srivastava, Geoffrey Hinton, Alex Krizhevsky, Ilya Sutskever, and Ruslan
  Salakhutdinov.
\newblock Dropout: a simple way to prevent neural networks from overfitting.
\newblock {\em The journal of machine learning research}, 15(1):1929--1958,
  2014.

\bibitem{sundararajan2017axiomatic}
Mukund Sundararajan, Ankur Taly, and Qiqi Yan.
\newblock Axiomatic attribution for deep networks.
\newblock In {\em International Conference on Machine Learning}, pages
  3319--3328. PMLR, 2017.

\bibitem{szegedy2017inception}
Christian Szegedy, Sergey Ioffe, Vincent Vanhoucke, and Alexander~A Alemi.
\newblock Inception-v4, inception-resnet and the impact of residual connections
  on learning.
\newblock In {\em Thirty-first AAAI conference on artificial intelligence},
  2017.

\bibitem{szegedy2016rethinking}
Christian Szegedy, Vincent Vanhoucke, Sergey Ioffe, Jon Shlens, and Zbigniew
  Wojna.
\newblock Rethinking the inception architecture for computer vision.
\newblock In {\em Proceedings of the IEEE conference on computer vision and
  pattern recognition}, pages 2818--2826, 2016.

\bibitem{szegedy2013intriguing}
Christian Szegedy, Wojciech Zaremba, Ilya Sutskever, Joan Bruna, Dumitru Erhan,
  Ian Goodfellow, and Rob Fergus.
\newblock Intriguing properties of neural networks.
\newblock {\em arXiv preprint arXiv:1312.6199}, 2013.

\bibitem{tramer2017ensemble}
Florian Tram{\`e}r, Alexey Kurakin, Nicolas Papernot, Ian Goodfellow, Dan
  Boneh, and Patrick McDaniel.
\newblock Ensemble adversarial training: Attacks and defenses.
\newblock {\em arXiv preprint arXiv:1705.07204}, 2017.

\bibitem{uesato2018adversarial}
Jonathan Uesato, Brendan O’donoghue, Pushmeet Kohli, and Aaron Oord.
\newblock Adversarial risk and the dangers of evaluating against weak attacks.
\newblock In {\em International Conference on Machine Learning}, pages
  5025--5034. PMLR, 2018.

\bibitem{wang2021feature}
Zhibo Wang, Hengchang Guo, Zhifei Zhang, Wenxin Liu, Zhan Qin, and Kui Ren.
\newblock Feature importance-aware transferable adversarial attacks.
\newblock In {\em Proceedings of the IEEE/CVF International Conference on
  Computer Vision}, pages 7639--7648, 2021.

\bibitem{wong2020fast}
Eric Wong, Leslie Rice, and J~Zico Kolter.
\newblock Fast is better than free: Revisiting adversarial training.
\newblock {\em arXiv preprint arXiv:2001.03994}, 2020.

\bibitem{wu2020boosting}
Weibin Wu, Yuxin Su, Xixian Chen, Shenglin Zhao, Irwin King, Michael~R Lyu, and
  Yu-Wing Tai.
\newblock Boosting the transferability of adversarial samples via attention.
\newblock In {\em Proceedings of the IEEE/CVF Conference on Computer Vision and
  Pattern Recognition}, pages 1161--1170, 2020.

\bibitem{wu2021improving}
Weibin Wu, Yuxin Su, Michael~R Lyu, and Irwin King.
\newblock Improving the transferability of adversarial samples with adversarial
  transformations.
\newblock In {\em Proceedings of the IEEE/CVF Conference on Computer Vision and
  Pattern Recognition}, pages 9024--9033, 2021.

\bibitem{xie2017mitigating}
Cihang Xie, Jianyu Wang, Zhishuai Zhang, Zhou Ren, and Alan Yuille.
\newblock Mitigating adversarial effects through randomization.
\newblock {\em arXiv preprint arXiv:1711.01991}, 2017.

\bibitem{xie2019improving}
Cihang Xie, Zhishuai Zhang, Yuyin Zhou, Song Bai, Jianyu Wang, Zhou Ren, and
  Alan~L Yuille.
\newblock Improving transferability of adversarial examples with input
  diversity.
\newblock In {\em Proceedings of the IEEE/CVF Conference on Computer Vision and
  Pattern Recognition}, pages 2730--2739, 2019.

\bibitem{zhou2018transferable}
Wen Zhou, Xin Hou, Yongjun Chen, Mengyun Tang, Xiangqi Huang, Xiang Gan, and
  Yong Yang.
\newblock Transferable adversarial perturbations.
\newblock In {\em Proceedings of the European Conference on Computer Vision
  (ECCV)}, pages 452--467, 2018.

\end{thebibliography}
}

\end{document}


\title{Improving Adversarial Transferability via Neuron Attribution-Based Attacks \\ Appendix}

\author{Jianping Zhang$ ^{1} $ \qquad
Weibin Wu$ ^{2} $\thanks{Corresponding author.} \qquad
Jen-tse Huang$ ^{1} $ \qquad
Yizhan Huang$ ^{1} $ \qquad
\\
Wenxuan Wang$ ^{1} $ \qquad
Yuxin Su$ ^{2} $ \qquad
Michael R. Lyu$ ^{1} $
\\
$ ^{1} $Department of Computer Science and Engineering, The Chinese University of Hong Kong
\\
$ ^{2} $School of Software Engineering, Sun Yat-sen University
\\
{\tt\small \{jpzhang, jthuang, yzhuang9, wxwang, lyu\}@cse.cuhk.edu.hk, \{wuwb36, suyx35\}@mail.sysu.edu.cn}
}

\maketitle

\renewcommand\thesection{\Alph{section}}
\renewcommand\thesubsection{\thesection.\arabic{subsection}}


The appendix includes two parts. Part \ref{e} discusses the potential negative impacts of this work on the society. Part \ref{f} rethinks the limitations of our approach.

\section{Potential Negative Societal Impacts}
\label{e}
Crafting adversarial examples has potential negative impacts on the society, because our approach can be misused by criminals to attack real-world systems. However, our work is important for figuring out the real internal defects of deep learning models. As a result, our work can motivate the community to design stronger defenses in the future.

\section{Limitations}
\label{f}
In our approach, we attack a single layer and conduct ablation studies to analyze attacking which layer could have the best transferability. Nevertheless, we have no guarantee that only attacking a single layer is the optimal strategy. It is possible that better results are achievable by attacking an ensemble of layers, but this would require tuning the layer combination. We leave it for future work.

\clearpage